\documentclass[conference]{IEEEtran}
\IEEEoverridecommandlockouts
\usepackage{cite}
\usepackage{amsmath,amssymb,amsfonts}
\usepackage{algorithmic}
\usepackage{graphicx}
\usepackage{textcomp}
\usepackage{xcolor}
\usepackage{booktabs}
\def\BibTeX{{\rm B\kern-.05em{\sc i\kern-.025em b}\kern-.08em
    T\kern-.1667em\lower.7ex\hbox{E}\kern-.125emX}}
\begin{document}

\title{FG-SGL: Fine-Grained Semantic Guidance Learning via Motion Process Decomposition for Micro-Gesture Recognition}

% \author{Jinsheng Wei, Zhaodi Xu, Guanming Lu, Haoyu Chen, Jingjie Yan}

\author{
Jinsheng Wei\textsuperscript{1}, Zhaodi Xu\textsuperscript{1}, Guanming Lu\textsuperscript{1}, Haoyu Chen\textsuperscript{2}, Jingjie Yan\textsuperscript{1} \\
\textsuperscript{1}Nanjing University of Posts and Telecommunications, China \\
\textsuperscript{2}University of Oulu, Finland
}

\maketitle

\begin{abstract}

Micro-gesture recognition (MGR) is challenging due to subtle inter-class variations.
Existing methods rely on category-level supervision, which is insufficient for capturing subtle and localized motion differences.
Thus, this paper proposes a Fine-Grained Semantic Guidance Learning (FG-SGL) framework that jointly integrates fine-grained and category-level semantics to guide vision--language models in perceiving local MG motions. FG-SA adopts fine-grained semantic cues to guide the learning of local motion features, while CP-A enhances the separability of MG features through category-level semantic guidance.
To support fine-grained semantic guidance, this work constructs a fine-grained textual dataset with human annotations that describes the dynamic process of MGs in four refined semantic dimensions. Furthermore, a Multi-Level Contrastive Optimization strategy is designed to jointly optimize both modules in a coarse-to-fine pattern. Experiments show that FG-SGL achieves competitive performance, validating the effectiveness of fine-grained semantic guidance for MGR.

\end{abstract}


\section{Introduction}
\label{sec:intro}

Micro-gestures (MGs) refer to subtle and involuntary body movements that occur within a very short duration, such as a brief touch to the neck or slight finger rubbing~\cite{chen2023smg,liu2021imigue}. These nonverbal behaviors often arise spontaneously during emotional fluctuations, increased cognitive load, or psychological stress. Due to their high level of authenticity and concealment, MGs can effectively reflect an individual’s underlying psychological state and emotional changes~\cite{pease1984body}. Compared to facial expressions or vocal signals, MG data are easier to collect and involve less privacy-sensitive information, making them particularly valuable in psychological assessment, intelligent human–computer interaction, and remote medical monitoring~\cite{krakovsky2018artificial}.

However, recognizing MGs from video remains highly challenging. First, the motion amplitude of MGs is extremely small, typically manifested as slight changes in localized regions, which can easily be overshadowed by overall appearance variations and background noise~\cite{shah2022efficient}. Second, the differences between different MG categories are often extremely subtle~\cite{gao2024identity}. For example, rubbing the eyes and touching the eyebrow exhibit highly similar hand trajectories, with category distinctions reflected only in pixel-level spatial displacements. Such minimal inter-class variability makes it difficult for models to extract stable and discriminative feature representations.

\begin{figure}[t]
\centering
\includegraphics[width=1\linewidth]{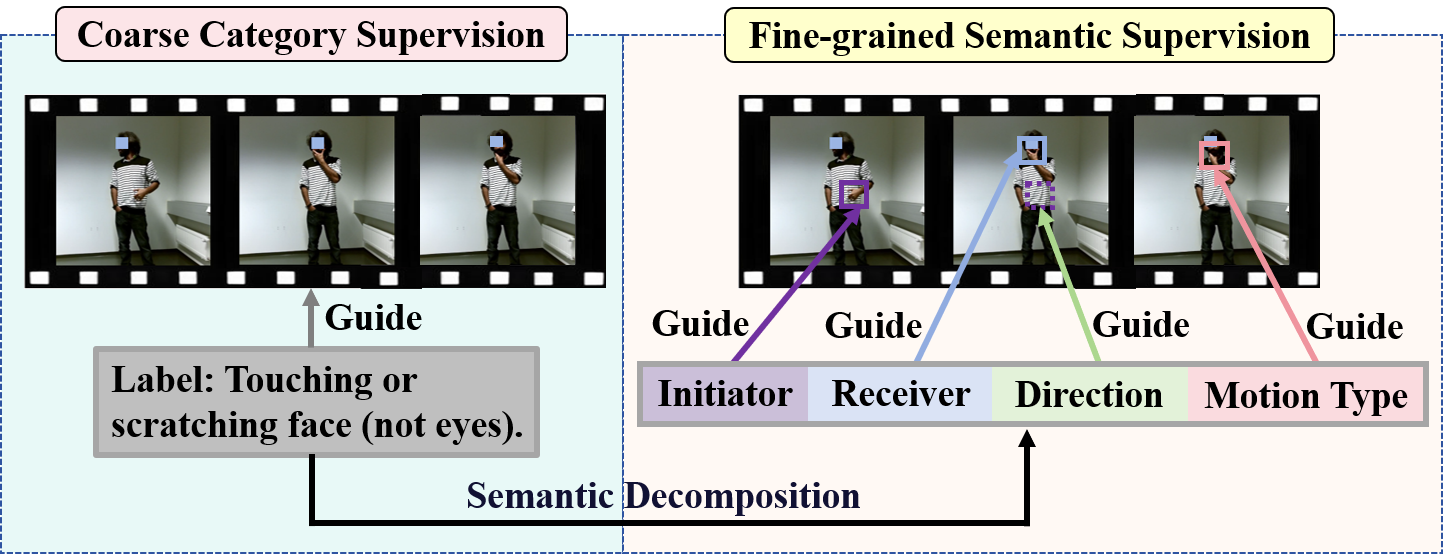}
\caption{Comparison between coarse category-level supervision and fine-grained semantic supervision for MGR. While coarse labels provide only global guidance, fine-grained semantics decompose a MG into localized motion attributes, enabling more discriminative supervision.}
\label{fig:motivation}
\end{figure}

% Existing methods~\cite{shah2024naive,chen2024prototype,li2023joint,xia2025hybrid,zhang20243d} for micro-gesture recognition (MGR) predominantly rely on category-level supervision during training. While such supervision specifies the target class of an action, it offers limited guidance on the fine-grained motion cues essential for distinguishing subtle differences between MG categories. As a result, models often focus on coarse dynamics for classification, overlooking the spatially localized features critical for reliable MGR. Considering that each action can be decomposed into finer semantic dimensions, we aim to integrate more precise motion cues into the recognition process. As shown in Fig.~\ref{fig:motivation}, fine-grained semantic supervision explicitly encodes motion information relevant to MGs. However, existing classification-based learning paradigms have yet to effectively incorporate such fine-grained semantic guidance into their representation learning.
Existing methods~\cite{shah2024naive,chen2024prototype,li2023joint,xia2025hybrid,zhang20243d} for micro-gesture recognition (MGR) predominantly rely on category-level supervision during training. While such supervision specifies the target class of an action, it offers limited guidance on the fine-grained motion cues essential for distinguishing subtle differences between MG categories. As a result, models often focus on coarse dynamics for classification, overlooking the spatially localized features critical for reliable MGR. In fact, each MG can be decomposed into finer semantic dimensions, such as the initiator, receiver, direction, and motion type, as shown in Fig.~\ref{fig:motivation}. Accordingly, this work integrates more precise motion cues into the recognition process.

However, existing classification-based learning paradigms have yet to effectively incorporate such fine-grained semantic guidance into their representation learning. Recent advances in vision--language models (VLMs)~\cite{radford2021learning,li2023videochat,liu2023visual,chen2024internvl,maaz2024video,wang2024qwen2}, particularly those trained with contrastive learning objectives, have demonstrated strong capability in aligning visual representations with semantic descriptions, offering a promising paradigm for semantic-guided representation learning. This paradigm provides a natural foundation for introducing fine-grained semantic supervision into MGR. Nevertheless, existing VLM-based approaches for MGR typically rely on coarse or holistic semantic descriptions, which remain insufficient for capturing the subtle and localized motion patterns inherent to MGs. In this work, we explore the effectiveness of fine-grained semantic guidance for MGR and propose a Fine-Grained Semantic Guidance Learning framework, termed FG-SGL, to systematically integrate fine-grained semantics into MG representation learning.

To support fine-grained semantic guidance learning, we construct FG-Text, a fine-grained textual dataset with human annotations that describes MGs by decomposing each instance into four refined semantic dimensions. By encoding localized motion semantics in natural language, FG-Text provides explicit semantic supervision beyond coarse category labels without altering the original category space. Furthermore, we introduce a unified Multi-Level Contrastive Optimization (ML-CO) strategy that jointly considers fine-grained semantic objectives and category-level objectives within a multi-level contrastive learning framework. ML-CO is designed to facilitate stable and effective learning of discriminative MG representations by coordinating semantic supervision at different representation levels.

In summary, this paper makes the following contributions:
\begin{itemize}
\item This work proposes a fine-grained semantic guidance learning framework for MGR, termed FG-SGL, which explicitly incorporates fine-grained semantic information into representation learning.
\item A fine-grained textual dataset, FG-Text, with human annotations is constructed to provide explicit semantic supervision for modeling localized motion semantics of MGs.
\item Two complementary modules, FG-SA and CP-A, are introduced to integrate fine-grained semantic guidance and category-level discrimination within a unified framework.
\item We propose a unified training strategy, termed ML-CO, which coordinates fine-grained and category-level semantic objectives through multi-level contrastive learning.
\end{itemize}

\section{Related work}
\subsection{Micro-gesture Datasets}
Recent progress in MG research has been supported by several benchmark datasets, including SMG~\cite{chen2023smg}, iMiGUE~\cite{liu2021imigue}, and MA-52~\cite{guo2024benchmarking}. SMG provides full-body RGB videos paired with skeleton sequences, while iMiGUE captures upper-body MGs in natural interview scenarios. MA-52 further expands the scale by offering large-scale RGB–skeleton aligned data with finer-grained category annotations. Although these datasets lay an important foundation for MGR, their annotations remain limited to coarse class labels and lack semantic descriptions of the internal action structure, which constrains the model’s ability to distinguish fine-grained motion differences.
\subsection{Micro-gesture Recognition}
Recent advances in MGR have explored different input modalities. Skeleton-based methods focus on modeling subtle motion dynamics from pose sequences. Gu et al.~\cite{gu2025motion} proposed Motion Matters, which amplifies fine temporal variations through motion-guided modulation. Xu et al.~\cite{xu2025towards} introduced topology-aware skeleton enhancement and temporal smoothing to obtain more stable skeleton representations. In the multimodal setting, CLIP-MG~\cite{patapati2025clip} leverages skeletal cues to guide semantic queries in a vision–language framework, and utilizes pose information to facilitate semantic alignment with RGB features. While these approaches improve MGR through structural modeling or multimodal fusion, they do not explicitly model the internal semantic structure of MGs. In contrast, this work introduces fine-grained semantic supervision from a vision--language perspective.

\section{Method}
\subsection{Overall Framework}

\begin{figure*}[t]
    \centering
    \includegraphics[width=\textwidth]{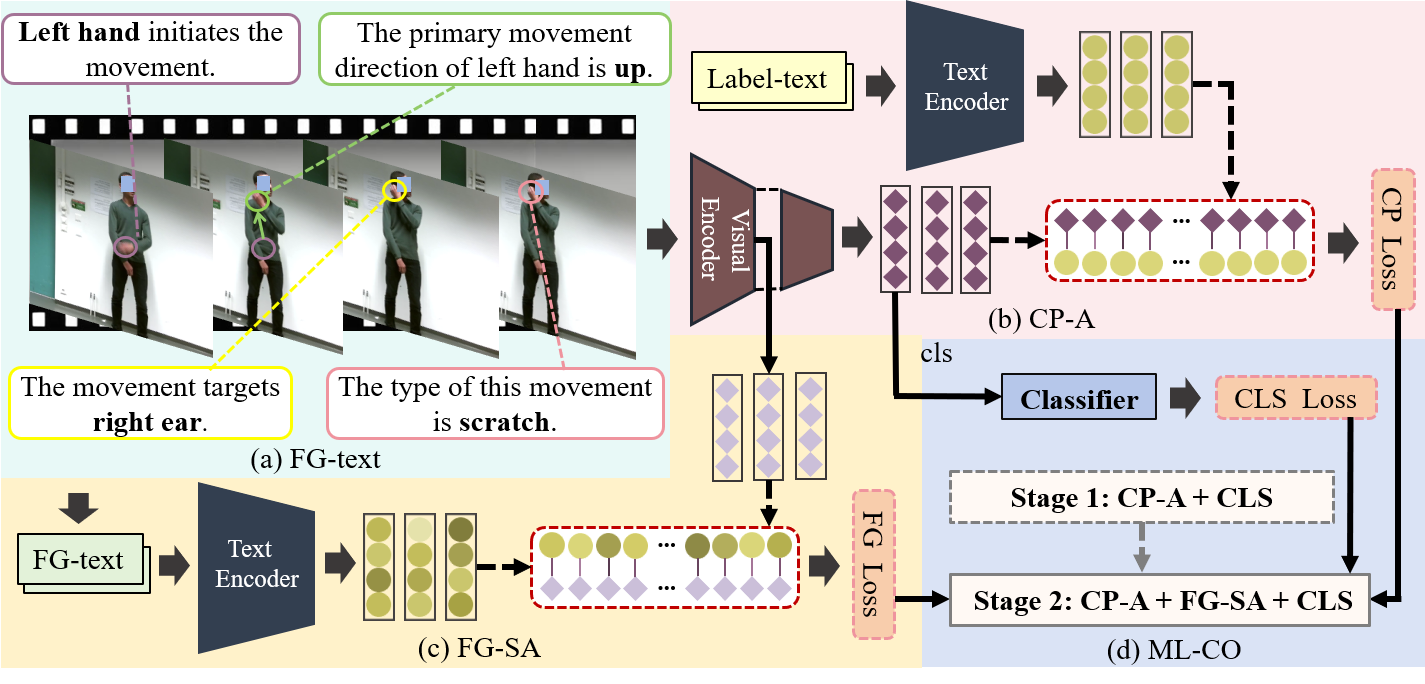}
    \caption{Overall framework of FG-SGL. The framework introduces fine-grained instance-aware semantic alignment (FG-SA) and category-level prototype alignment (CP-A) on mid-level and high-level video representations, respectively, and jointly optimizes them under a unified learning objective.}
    \label{figframework}
\end{figure*}

FG-SGL is a fine-grained semantic guidance learning framework for MGR, built upon a pretrained vision--language model. As shown in Fig.~\ref{figframework}, the framework integrates multiple semantic signals into the visual representation learning process through a set of dedicated modules.

Given an input video $V$, the visual encoder produces hierarchical visual representations. FG-SGL makes use of representations from different depths of the visual encoder, which are associated with semantic signals of different granularities. In parallel, textual descriptions at both instance and category levels are encoded by the text encoder and serve as semantic references.

The framework consists of three key components: a structured semantic prior (FG-Text), fine-grained semantic alignment (FG-SA), and category prototype alignment (CP-A). These components interact with visual representations at different levels and are jointly integrated within a unified learning framework. The detailed designs of each component are introduced in the following subsections.

\subsection{Structured Semantic Prior: FG-Text}

To provide fine-grained semantic supervision for MGR, we introduce FG-Text, a structured semantic prior that provides instance-aware semantic descriptions for each video clip. Unlike coarse category labels, FG-Text captures localized and interpretable semantic cues that characterize how a MG is performed.

Specifically, FG-Text decomposes a MG into four semantic attributes: \emph{initiator}, \emph{receiver}, \emph{direction}, and \emph{motion type}. These attributes describe which body part initiates the motion, which target is involved, the primary movement direction, and the motion pattern, respectively. Together, they form a structured semantic description that reflects the internal action structure of a MG.

For each video sample, the four semantic attributes are manually annotated and rigorously cross-validated, and then combined into a natural-language description, denoted as $T_{fg}$. This description is encoded by the text encoder to obtain an instance-aware semantic embedding, which serves as a fine-grained semantic reference in the subsequent alignment process.

FG-Text functions as a semantic prior that complements visual representations by providing explicit and localized semantic information. It enables the framework to associate visual motion patterns with interpretable semantic attributes, facilitating fine-grained semantic grounding at the instance level.

\subsection{Fine-Grained Semantic Alignment (FG-SA)}

Fine-Grained Semantic Alignment (FG-SA) aims to associate localized visual motion patterns with instance-aware fine-grained semantics. Given that subtle MG variations are often reflected in spatially and temporally localized regions, FG-SA operates on mid-level visual representations, which preserve detailed motion information while retaining sufficient semantic abstraction.

Specifically, for an input video, the visual encoder produces a mid-level representation $\mathbf{f}^{mid}$. In parallel, the instance-aware fine-grained semantic description $T_{fg}$ is encoded by the text encoder to obtain a semantic embedding $\mathbf{t}_{fg}$. FG-SA encourages alignment between $\mathbf{f}^{mid}$ and $\mathbf{t}_{fg}$ by bringing semantically corresponding visual and textual representations closer in the shared embedding space.

To achieve this, we adopt a contrastive learning objective that treats each video--text pair as a positive sample, while other semantic descriptions within the same batch serve as negatives. The fine-grained semantic alignment loss is formulated as:
\begin{equation}
\mathcal{L}_{fg} =
- \log
\frac{
\exp\!\left( \mathrm{sim}\!\left( \mathbf{f}_{\mathrm{mid}}, \mathbf{t}_{fg} \right) / \tau \right)
}{
\sum_{j}
\exp\!\left( \mathrm{sim}\!\left( \mathbf{f}_{\mathrm{mid}}, \mathbf{t}_{fg}^{(j)} \right) / \tau \right)
},
\end{equation}
where $\mathrm{sim}(\cdot)$ denotes cosine similarity and $\tau$ is a temperature parameter.

By aligning mid-level visual representations with fine-grained semantic embeddings, FG-SA encourages the model to ground localized motion cues in explicit semantic attributes, thereby enhancing its sensitivity to subtle motion variations among visually similar MG instances.

\subsection{Category Prototype Alignment (CP-A)}

While FG-SA focuses on grounding localized motion patterns at the instance level, MGR also requires robust discrimination across MG categories. To this end, we introduce Category Prototype Alignment (CP-A), which imposes semantic structure at the category level by aligning high-level visual representations with category-level textual semantics.

For an input video, the visual encoder produces a high-level visual representation $\mathbf{f}^{high}$ that encodes the global semantics of the entire gesture. In parallel, each gesture category is associated with a textual description $T_c$, which is encoded by the text encoder to obtain a category-level semantic embedding $\mathbf{t}_c$. CP-A aligns $\mathbf{f}^{high}$ with the corresponding category semantic embedding, encouraging samples from the same class to be mapped close to their shared semantic prototype.

Specifically, a contrastive learning objective is adopted at the category level, where the matched video--category pair is treated as a positive sample and other category descriptions serve as negatives:
\begin{equation}
\mathcal{L}_{cp} =
- \log
\frac{
\exp\!\left( \mathrm{sim}\!\left( \mathbf{f}_{\mathrm{high}}, \mathbf{t}_{c} \right) / \tau \right)
}{
\sum_{k}
\exp\!\left( \mathrm{sim}\!\left( \mathbf{f}_{\mathrm{high}}, \mathbf{t}_{c}^{(k)} \right) / \tau \right)
},
\end{equation}
where $\mathrm{sim}(\cdot)$ denotes cosine similarity and $\tau$ is a temperature parameter.

By aligning high-level visual representations with category-level semantic prototypes, CP-A enforces global semantic consistency and enhances inter-class separability among visually similar MG categories.

\subsection{Multi-Level Contrastive Optimization}

To effectively integrate category-level and fine-grained semantic supervision, we adopt a progressive training strategy termed Multi-Level Contrastive Optimization (ML-CO). ML-CO is designed to coordinate contrastive objectives defined at different representation levels, enabling stable optimization of fine-grained semantic alignment.

Specifically, ML-CO consists of two training stages. In the first stage, the model is optimized using the classification loss $\mathcal{L}_{cls}$ together with the category prototype alignment loss $\mathcal{L}_{cp}$. This stage aims to establish a discriminative and semantically consistent category-level representation by aligning high-level visual features with category prototypes.

In the second stage, the fine-grained semantic alignment loss $\mathcal{L}_{fg}$ is further introduced, while $\mathcal{L}_{cls}$ and $\mathcal{L}_{cp}$ remain active. By incorporating instance-aware fine-grained semantic supervision at this stage, the model refines mid-level visual representations to capture localized motion patterns within each category.

Throughout both stages, the classification loss $\mathcal{L}_{cls}$ is applied to enforce task-level supervision and maintain discriminative decision boundaries. The overall training objective is formulated as:
\begin{equation}
\mathcal{L}
=
\mathcal{L}_{cls}
+ \lambda_{fg}\mathcal{L}_{fg}
+ \lambda_{cp}\mathcal{L}_{cp},
\end{equation}
where $\lambda_{fg}$ and $\lambda_{cp}$ control the contributions of fine-grained and category-level contrastive objectives, respectively.

By progressively introducing semantic supervision at multiple levels and jointly optimizing all objectives, ML-CO enables FG-SGL to capture subtle motion variations while maintaining robust category-level discrimination within a unified learning framework.

\section{Experiments}
\subsection{Experimental Setup}

\paragraph{Datasets and protocol}
We evaluate the proposed FG-SGL framework on two widely used MG benchmarks, SMG~\cite{chen2023smg} and iMiGUE~\cite{liu2021imigue}. 
For SMG, we follow the official data split, using 2{,}452 video clips for training, 637 clips for validation, and 610 clips for testing. 
The model is trained on the training split, and the checkpoint achieving the highest top-1 accuracy on the validation set is selected for final evaluation, with all reported results obtained on the held-out test set. 
For iMiGUE, we adopt the standard evaluation protocol provided by the dataset and report top-1 classification accuracy on the test split.
For the fine-grained semantic supervision used in our experiments, annotation consistency is verified via cross-validation among annotators to ensure reliable semantic guidance.

\paragraph{Video preprocessing}
For both datasets, we uniformly sample 8 frames from each video clip and resize them to a spatial resolution of $448 \times 448$ before feeding them into the visual encoder.

\paragraph{Backbone and model configuration}
FG-SGL is instantiated on InternVL2.5-8B~\cite{chen2024internvl}, which provides a ViT-based visual encoder and a transformer-based text encoder. 
The text encoder is kept frozen throughout training. 
On the visual side, we freeze the backbone weights and insert LoRA adapters with rank 8 into the vision encoder to enable parameter-efficient fine-tuning, while the MLP layers are fully trainable. 
Unless otherwise specified, the same model configuration is used across all experiments.

\paragraph{Training details}
All experiments are implemented with HuggingFace Transformers and DeepSpeed. 
We use AdamW with a cosine learning rate schedule and a warm-up ratio of 0.03. 
The learning rate is set to $4 \times 10^{-5}$ for SMG and $1 \times 10^{-4}$ for iMiGUE, with weight decay 0.05 and gradient clipping at 1.0. 
Models are trained for 15 epochs using {bf16 precision on two NVIDIA A100 GPUs. 
The per-device batch size is 8 with 4-step gradient accumulation, and the checkpoint with the highest validation top-1 accuracy is selected.
All models are trained following the two-stage ML-CO strategy described in Sec.~III-E, where category-level alignment is optimized first, followed by joint optimization with fine-grained semantic alignment.

\subsection{Baselines}

The proposed FG-SGL is evaluated against representative MGR methods spanning three major paradigms: RGB-based models, skeleton-based models, and multimodal RGB–skeleton approaches.

\textbf{RGB-based methods.}
As RGB-only baselines, we include TSM~\cite{lin2019tsm}, a commonly adopted video recognition backbone, as well as CVTCL~\cite{li2024enhancing} that leverages category-level textual descriptions during training. These approaches rely exclusively on RGB appearance and temporal information.

\textbf{Skeleton-based methods.}
Skeleton-driven MGR is represented by methods such as JSSEL~\cite{li2023joint} and H2OFormer~\cite{xia2025hybrid}, which model fine-grained body dynamics from pose sequences and achieve strong performance using skeleton data alone.

\textbf{Multimodal RGB–Skeleton methods.}
In addition, we include representative multimodal approaches, including PL~\cite{chen2024prototype} and M2HEN~\cite{huang2024multi}, which combine RGB appearance with pose-based motion cues for MGR. These methods provide reference points for evaluating performance under multimodal inputs.

Overall, these baselines cover the main methodological directions in existing MGR research.

\subsection{Comparison with Existing Methods}

Table~\ref{tab:main_results} reports the MGR results of different methods on SMG and iMiGUE.
On SMG, FG-SGL achieves the best performance compared to other methods. Compared with RGB-based and skeleton-based baselines, FG-SGL demonstrates improved recognition accuracy, indicating that incorporating structured semantic guidance effectively benefits fine-grained MGR. This improvement highlights the ability of FG-SGL to capture subtle, localized motion differences that are critical for distinguishing between similar MGs, which traditional methods might miss.

On iMiGUE, FG-SGL outperforms CLIP-based baselines and achieves competitive performance among RGB-only methods. Despite relying solely on RGB videos at inference time, FG-SGL benefits from structured semantic supervision, which partially compensates for the absence of explicit pose information and enhances the discriminative capability of RGB-based representations. While some skeleton-based and multimodal approaches still achieve higher accuracy, FG-SGL offers a strong and effective complement, particularly in settings where fine-grained motion discrimination is essential.

Taken together, the results on both datasets indicate that aligning visual features with structured textual semantics yields a consistent performance advantage.

\begin{table}[t]
\caption{Comparison of different methods on SMG and iMiGUE. Top-1 accuracy (\%) is reported.}
\begin{center}
\begin{tabular}{|l|c|c|c|c|}
\hline
\textbf{Method} & \textbf{Year} & \textbf{Modality} & \textbf{SMG} & \textbf{iMiGUE} \\
\hline
TSM~\cite{lin2019tsm} & 2019 & RGB & 65.41 & N/A \\
\hline
MS-G3D~\cite{liu2020disentangling} & 2020 & Skeleton & 64.75 & N/A \\
\hline
JSSEL~\cite{li2023joint} & 2023 & Skeleton & 68.03 & 64.12 \\
\hline
CVTCL\cite{li2024enhancing} & 2024 & RGB & 65.08 & 66.12 \\
\hline
MMGC~\cite{wang2024multimodal} & 2024 & Skeleton & N/A & 68.90 \\
\hline
PL~\cite{chen2024prototype} & 2024 & RGB + Skeleton & N/A & 70.25 \\
\hline
M2HEN~\cite{huang2024multi} & 2024 & RGB + Skeleton & N/A & 70.19 \\
\hline
H2OFormer~\cite{xia2025hybrid} & 2025 & Skeleton & 75.44 & 70.00 \\
\hline
MMN~\cite{gu2025motion} & 2025 & Skeleton & N/A & 60.21 \\
\hline
CLIP-MG~\cite{patapati2025clip} & 2025 & RGB + Skeleton & N/A & 61.82 \\
\hline
\textbf{FG-SGL (Ours)} & \textbf{2025} & \textbf{RGB} & \textbf{78.13} & \textbf{62.58} \\
\hline
\end{tabular}
\label{tab:main_results}
\end{center}
\end{table}

\subsection{Ablation Studies}

\begin{table}[t]
\caption{Ablation study of different components in FG-SGL on SMG and iMiGUE. Top-1 accuracy (\%) is reported.}
\label{tab:ablation_main}
\begin{center}
\begin{tabular}{|c|c|c|c|c|}
\hline
\textbf{FG-SA} & \textbf{CP-A} & \textbf{ML-CO} & \textbf{SMG} & \textbf{iMiGUE} \\
\hline
$\times$ & $\checkmark$ & $\checkmark$ & 76.51 & 59.29 \\
\hline
$\checkmark$ & $\times$ & $\checkmark$ & 73.22 & 57.08 \\
\hline
$\checkmark$ & $\checkmark$ & $\times$ & 77.70 & 60.72 \\
\hline
$\checkmark$ & $\checkmark$ & $\checkmark$ & \textbf{78.13} & \textbf{62.58} \\
\hline
\end{tabular}
\end{center}
\end{table}

We evaluate the contribution of each component in FG-SGL through ablation studies on SMG and iMiGUE.
All variants share the same backbone architecture and training configuration, differing only in whether fine-grained semantic alignment (FG-SA), category prototype alignment (CP-A), and multi-level contrastive optimization (ML-CO) are applied.

\paragraph{Effect of Fine-Grained Semantic Alignment (FG-SA)}
Table~\ref{tab:ablation_main} reports the effect of fine-grained semantic alignment (FG-SA) on recognition performance.
When FG-SA is removed while keeping CP-A and ML-CO unchanged, the recognition accuracy decreases from 78.13\% to 76.51\% on SMG and from 62.58\% to 59.29\% on iMiGUE. This performance drop suggests that fine-grained semantic alignment significantly contributes to the model’s ability to capture subtle and localized motion patterns in MGR. Without FG-SA, the model has to rely solely on coarse category-level information, which is insufficient to differentiate the highly similar motion patterns found in MGs.
The larger performance degradation observed on iMiGUE indicates that fine-grained alignment becomes even more crucial when dealing with datasets that contain more complex or nuanced motion patterns, such as those in iMiGUE. The higher inter-class similarity in iMiGUE, combined with subtle motion variations, makes it difficult for the model to differentiate between categories without fine-grained guidance.

\paragraph{Effect of Category Prototype Alignment (CP-A)}
Table~\ref{tab:ablation_main} also presents the impact of category prototype alignment (CP-A) on model performance.
When CP-A is removed while keeping FG-SA and ML-CO enabled, the recognition accuracy drops from 78.13\% to 73.22\% on SMG and from 62.58\% to 57.08\% on iMiGUE. This significant performance drop emphasizes the importance of category-level semantic alignment in maintaining global discriminability among MG categories.
CP-A plays a crucial role in reinforcing the separation between visually similar categories. Without CP-A, the model struggles to maintain distinct class boundaries, leading to increased inter-class confusion. For instance, in scenarios where the difference between categories is subtle but still present (e.g., rubbing eyes vs. touching the eyebrow), CP-A ensures that the model can better separate these categories based on category-level semantic prototypes. This separation is particularly essential when handling complex datasets like iMiGUE, where subtle body movements occur across multiple categories. Without CP-A, the model's capacity to discriminate between such categories diminishes, resulting in lower recognition accuracy.

% \paragraph{Effect of Multi-Level Contrastive Optimization (ML-CO)}
% The effect of the proposed ML-CO strategy is reported in Table~II.
% When ML-CO is disabled and all objectives are jointly optimized from the beginning of training, the recognition accuracy decreases from 78.13\% to 77.70\% on SMG and from 62.58\% to 60.72\% on iMiGUE.
% Unlike FG-SA and CP-A, ML-CO does not introduce an additional supervision signal, but instead changes the optimization manner by progressively introducing fine-grained semantic alignment after category-level representations have been established.
% This comparison indicates that directly optimizing all contrastive objectives from scratch can lead to suboptimal alignment, especially for instance-aware fine-grained semantics.
% In contrast, the progressive training strategy in ML-CO provides a more stable optimization process, resulting in clearer performance gains on the more complex iMiGUE dataset.

\paragraph{Effect of Multi-Level Contrastive Optimization (ML-CO)}
The effect of the proposed ML-CO strategy is reported in Table~\ref{tab:ablation_main}.
When ML-CO is disabled and all objectives are jointly optimized from the beginning of training, the recognition accuracy decreases from 78.13\% to 77.70\% on SMG and from 62.58\% to 60.72\% on iMiGUE.
ML-CO modifies the optimization dynamics by progressively introducing fine-grained semantic alignment after category-level representations have been established.
Directly optimizing both category-level and instance-aware contrastive objectives from scratch can lead to conflicting gradients at early training stages, where visual representations are not yet well structured.
As a result, fine-grained semantic alignment becomes less stable and less effective.
By first enforcing category-level semantic consistency, ML-CO provides a more organized representation space, enabling fine-grained alignment to be performed more reliably.
This progressive optimization strategy results in more stable training and yields clearer performance gains, particularly on the more challenging iMiGUE dataset.

\begin{table}[t]
\caption{Ablation study on the design of FG-Text. Top-1 accuracy (\%) is reported.}
\label{tab:ablation_fgtext}
\begin{center}
\begin{tabular}{|c|c|c|c|c|c|}
\hline
\textbf{FG-SA} & \textbf{CP-A} & \textbf{ML-CO} & \textbf{FG-Text} & \textbf{SMG} & \textbf{iMiGUE} \\
\hline
$\checkmark$ & $\checkmark$ & $\checkmark$ & Class-level & 76.55 & 59.36 \\
\hline
$\checkmark$ & $\checkmark$ & $\checkmark$ & Fine-grained & \textbf{78.13} & \textbf{62.58} \\
\hline
\end{tabular}
\end{center}
\end{table}

\paragraph{Ablation on FG-Text}
Table~\ref{tab:ablation_fgtext} reports the effect of replacing fine-grained FG-Text with Class-level text. Class-level text refers to a single holistic textual description shared by all video clips of the same gesture category, which encodes only coarse category semantics without instance-aware fine-grained attributes.
When FG-Text is substituted by class-level text while keeping all other components unchanged, the recognition accuracy decreases from 78.13\% to 76.55\% on SMG and from 62.58\% to 59.36\% on iMiGUE.
This performance drop indicates that category-level descriptions alone are insufficient to provide effective supervision for fine-grained semantic alignment.
In particular, the larger degradation observed on iMiGUE suggests that modeling instance-aware motion semantics becomes more important in datasets with higher inter-class similarity and more diverse motion patterns.

\section{Conclusion}
In this paper, we investigated the role of semantic supervision in MGR and proposed FG-SGL, a fine-grained semantic guidance learning framework that integrates structured textual semantics into visual representation learning at multiple levels.
By aligning mid-level visual features with fine-grained motion semantics and high-level representations with category-level descriptions, FG-SGL effectively enhances the discriminability of visually similar MG categories.
The introduction of FG-Text provides explicit semantic decomposition of MG motions. Moreover, the proposed multi-level contrastive optimization strategy adopts a coarse-to-fine training scheme, where category-level alignment is first established to stabilize the representation space before introducing fine-grained semantic supervision.

Experimental results on SMG and iMiGUE demonstrate that incorporating structured semantic guidance significantly improves MGR performance.
In future work, we plan to explore more scalable semantic representations and extend the proposed framework to broader fine-grained action understanding tasks.

\bibliographystyle{IEEEbib}
\bibliography{icme2026references}

@article{chen2023smg,
  title={Smg: A micro-gesture dataset towards spontaneous body gestures for emotional stress state analysis},
  author={Chen, Haoyu and Shi, Henglin and Liu, Xin and Li, Xiaobai and Zhao, Guoying},
  journal={International Journal of Computer Vision},
  volume={131},
  number={6},
  pages={1346--1366},
  year={2023},
  publisher={Springer}
}

@inproceedings{liu2021imigue,
  title={imigue: An identity-free video dataset for micro-gesture understanding and emotion analysis},
  author={Liu, Xin and Shi, Henglin and Chen, Haoyu and Yu, Zitong and Li, Xiaobai and Zhao, Guoying},
  booktitle={Proceedings of the IEEE/CVF conference on computer vision and pattern recognition},
  pages={10631--10642},
  year={2021}
}

@article{krakovsky2018artificial,
  title={Artificial (emotional) intelligence},
  author={Krakovsky, Marina},
  journal={Communications of the ACM},
  volume={61},
  number={4},
  pages={18--19},
  year={2018},
  publisher={ACM New York, NY, USA}
}

@book{pease1984body,
  title={Body Language How to read others’ thoughts by their gestures},
  author={Pease, Allan},
  year={1984},
  publisher={Sheldon press}
}

@inproceedings{shah2022efficient,
  title={Efficient dense-graph convolutional network with inductive prior augmentations for unsupervised micro-gesture recognition},
  author={Shah, Atif and Chen, Haoyu and Shi, Henglin and Zhao, Guoying},
  booktitle={2022 26th International Conference on Pattern Recognition (ICPR)},
  pages={2686--2692},
  year={2022},
  organization={IEEE}
}

@article{guo2024benchmarking,
  title={Benchmarking micro-action recognition: Dataset, methods, and applications},
  author={Guo, Dan and Li, Kun and Hu, Bin and Zhang, Yan and Wang, Meng},
  journal={IEEE Transactions on Circuits and Systems for Video Technology},
  volume={34},
  number={7},
  pages={6238--6252},
  year={2024},
  publisher={IEEE}
}

@inproceedings{gu2025motion,
  title={Motion matters: Motion-guided modulation network for skeleton-based micro-action recognition},
  author={Gu, Jihao and Li, Kun and Wang, Fei and Wei, Yanyan and Wu, Zhiliang and Fan, Hehe and Wang, Meng},
  booktitle={Proceedings of the 33rd ACM International Conference on Multimedia},
  pages={5461--5470},
  year={2025}
}

@article{xu2025towards,
  title={Towards Fine-Grained Emotion Understanding via Skeleton-Based Micro-Gesture Recognition},
  author={Xu, Hao and Cheng, Lechao and Wang, Yaxiong and Tang, Shengeng and Zhong, Zhun},
  journal={arXiv preprint arXiv:2506.12848},
  year={2025}
}

@article{patapati2025clip,
  title={CLIP-MG: Guiding Semantic Attention with Skeletal Pose Features and RGB Data for Micro-Gesture Recognition on the iMiGUE Dataset},
  author={Patapati, Santosh and Srinivasan, Trisanth and Adiraju, Amith},
  journal={arXiv preprint arXiv:2506.16385},
  year={2025}
}

@inproceedings{shah2024naive,
  title={Naive data augmentation might be toxic: Data-prior guided self-supervised representation learning for micro-gesture recognition},
  author={Shah, Atif and Chen, Haoyu and Zhao, Guoying},
  booktitle={2024 IEEE 18th International Conference on Automatic Face and Gesture Recognition (FG)},
  pages={1--9},
  year={2024},
  organization={IEEE}
}

@article{chen2024prototype,
  title={Prototype learning for micro-gesture classification},
  author={Chen, Guoliang and Wang, Fei and Li, Kun and Wu, Zhiliang and Fan, Hehe and Yang, Yi and Wang, Meng and Guo, Dan},
  journal={arXiv preprint arXiv:2408.03097},
  year={2024}
}

@article{li2023joint,
  title={Joint skeletal and semantic embedding loss for micro-gesture classification},
  author={Li, Kun and Guo, Dan and Chen, Guoliang and Peng, Xinge and Wang, Meng},
  journal={arXiv preprint arXiv:2307.10624},
  year={2023}
}

@article{xia2025hybrid,
  title={Hybrid-supervised Hypergraph-enhanced Transformer for Micro-gesture Based Emotion Recognition},
  author={Xia, Zhaoqiang and Huang, Hexiang and Chen, Haoyu and Feng, Xiaoyi and Zhao, Guoying},
  journal={IEEE Transactions on Affective Computing},
  year={2025},
  publisher={IEEE}
}

@inproceedings{zhang20243d,
  title={3D Convolutional Network based micro-gesture recognition},
  author={Zhang, Congyue and Fu, Wenjie and Tian, Canrong and Cheng, Xu and Tian, Yuan and Yu, Hao},
  booktitle={Proceedings of the ACM Turing Award Celebration Conference-China 2024},
  pages={193--198},
  year={2024}
}

@article{li2023videochat,
  title={Videochat: Chat-centric video understanding},
  author={Li, KunChang and He, Yinan and Wang, Yi and Li, Yizhuo and Wang, Wenhai and Luo, Ping and Wang, Yali and Wang, Limin and Qiao, Yu},
  journal={arXiv preprint arXiv:2305.06355},
  year={2023}
}

@article{liu2023visual,
  title={Visual instruction tuning},
  author={Liu, Haotian and Li, Chunyuan and Wu, Qingyang and Lee, Yong Jae},
  journal={Advances in neural information processing systems},
  volume={36},
  pages={34892--34916},
  year={2023}
}

@inproceedings{maaz2024video,
  title={Video-chatgpt: Towards detailed video understanding via large vision and language models},
  author={Maaz, Muhammad and Rasheed, Hanoona and Khan, Salman and Khan, Fahad},
  booktitle={Proceedings of the 62nd Annual Meeting of the Association for Computational Linguistics (Volume 1: Long Papers)},
  pages={12585--12602},
  year={2024}
}

@article{wang2024qwen2,
  title={Qwen2-vl: Enhancing vision-language model's perception of the world at any resolution},
  author={Wang, Peng and Bai, Shuai and Tan, Sinan and Wang, Shijie and Fan, Zhihao and Bai, Jinze and Chen, Keqin and Liu, Xuejing and Wang, Jialin and Ge, Wenbin and others},
  journal={arXiv preprint arXiv:2409.12191},
  year={2024}
}

@inproceedings{radford2021learning,
  title={Learning transferable visual models from natural language supervision},
  author={Radford, Alec and Kim, Jong Wook and Hallacy, Chris and Ramesh, Aditya and Goh, Gabriel and Agarwal, Sandhini and Sastry, Girish and Askell, Amanda and Mishkin, Pamela and Clark, Jack and others},
  booktitle={International conference on machine learning},
  pages={8748--8763},
  year={2021},
  organization={PmLR}
}

@inproceedings{chen2024internvl,
  title={Internvl: Scaling up vision foundation models and aligning for generic visual-linguistic tasks},
  author={Chen, Zhe and Wu, Jiannan and Wang, Wenhai and Su, Weijie and Chen, Guo and Xing, Sen and Zhong, Muyan and Zhang, Qinglong and Zhu, Xizhou and Lu, Lewei and others},
  booktitle={Proceedings of the IEEE/CVF conference on computer vision and pattern recognition},
  pages={24185--24198},
  year={2024}
}

@article{gao2024identity,
  title={Identity-free artificial emotional intelligence via micro-gesture understanding},
  author={Gao, Rong and Liu, Xin and Xing, Bohao and Yu, Zitong and Schuller, Bjorn W and K{\"a}lvi{\"a}inen, Heikki},
  journal={arXiv preprint arXiv:2405.13206},
  year={2024}
}

@inproceedings{lin2019tsm,
  title={Tsm: Temporal shift module for efficient video understanding},
  author={Lin, Ji and Gan, Chuang and Han, Song},
  booktitle={Proceedings of the IEEE/CVF international conference on computer vision},
  pages={7083--7093},
  year={2019}
}

@article{li2024enhancing,
  title={Enhancing micro gesture recognition for emotion understanding via context-aware visual-text contrastive learning},
  author={Li, Deng and Xing, Bohao and Liu, Xin},
  journal={IEEE Signal Processing Letters},
  volume={31},
  pages={1309--1313},
  year={2024},
  publisher={IEEE}
}

@article{huang2024multi,
  title={Multi-modal micro-gesture classification via multiscale heterogeneous ensemble network},
  author={Huang, Hexiang and Wang, Yuhan and Kerui, L and Xia, Zhaoqiang},
  journal={MiGA@ IJCAI},
  year={2024}
}

@inproceedings{liu2020disentangling,
  title={Disentangling and unifying graph convolutions for skeleton-based action recognition},
  author={Liu, Ziyu and Zhang, Hongwen and Chen, Zhenghao and Wang, Zhiyong and Ouyang, Wanli},
  booktitle={Proceedings of the IEEE/CVF conference on computer vision and pattern recognition},
  pages={143--152},
  year={2020}
}

@misc{wang2024multimodal,
  title={A Multimodal Micro-gesture Classification Model Based on CLIP},
  author={Wang, Yiwen and Dong, Zhenyang and Li, Pengxia and Liu, Yujie},
  year={2024},
  publisher={MiGA}
}

\vspace{12pt}
\color{red}
% IEEE conference templates contain guidance text for composing and formatting conference papers. Please ensure that all template text is removed from your conference paper prior to submission to the conference. Failure to remove the template text from your paper may result in your paper not being published.

\end{document}